\pdfoutput=1
\documentclass{article} %
\usepackage{iclr2016_conference,times}
\usepackage[utf8]{inputenc} 
\usepackage{amsmath, amssymb, bm, mathtools, thmtools}
\usepackage{verbatim}
\usepackage{graphicx}\graphicspath{{figures/}}
\usepackage{multicol}
\usepackage{tabularx}
\usepackage{mathrsfs} 
\usepackage{url}

\usepackage[colorlinks,citecolor=blue,urlcolor=blue,linkcolor=blue]{hyperref}
\usepackage{hypernat}
\usepackage{datetime}

\DeclareMathAlphabet\EuRoman{U}{eur}{m}{n}
\SetMathAlphabet\EuRoman{bold}{U}{eur}{b}{n}

\usepackage{euscript,microtype}
\usepackage[capitalize]{cleveref}

\crefname{lemma}{Lemma}{Lemmas}
\crefname{corollary}{Corollary}{Corollaries}
\crefname{theorem}{Theorem}{Theorems}

\makeatletter
\let\reftagform@=\tagform@
\def\tagform@#1{\maketag@@@{\ignorespaces\textcolor{gray}{(#1)}\unskip\@@italiccorr}}
\renewcommand{\eqref}[1]{\textup{\reftagform@{\ref{#1}}}}
\makeatother

\newcommand{\LATER}[1]{\error}
\newcommand{\fLATER}[1]{\error}
\newcommand{\TBD}[1]{\error}
\newcommand{\fTBD}[1]{}
\newcommand{\PROBLEM}[1]{\error}
\newcommand{\fPROBLEM}[1]{}
\newcommand{\NA}[1]{#1}

\def\[#1\]{\begin{align}#1\end{align}}
\def\*[#1\]{\begin{align*}#1\end{align*}}

\newcommand{\defas}{\vcentcolon=}  %

\newcommand{\Reals}{\mathbb{R}}

\DeclareMathOperator*{\newlim}{\mathrm{lim}\vphantom{\mathrm{infsup}}}

\renewcommand{\lim}{\newlim}

\usepackage{times,caption}

\title{Neural Network Matrix Factorization}

\author{Gintare Karolina Dziugaite \\
Department of Engineering \\
University of Cambridge\\
Cambridge, United Kingdom CB2 1PZ \\
\texttt{gkd22@cam.ac.uk} \\
\And
Daniel M. Roy \\
Department of Statistical Sciences \\
University of Toronto\\
Toronto, Canada M5S 3G3\\
\texttt{droy@utstat.toronto.edu}
}

\newcommand{\hprod}{\circ}

\begin{document}

\maketitle

\begin{abstract}
Data often comes in the form of an array or matrix.  Matrix factorization techniques attempt to recover missing or corrupted entries by assuming that the matrix can be written as the product of two low-rank matrices.  In other words, matrix factorization approximates the entries of the matrix by a simple, fixed function---namely, the inner product---acting on the latent feature vectors for the corresponding row and column.  Here we consider replacing the inner product by an arbitrary function that we learn from the data at the same time as we learn the latent feature vectors.  In particular, we replace the inner product by a multi-layer feed-forward neural network, and learn by alternating between optimizing the network for fixed latent features, and optimizing the latent features for a fixed network.  
The resulting approach---which we call \emph{neural network matrix factorization} or NNMF, for short---dominates standard low-rank techniques on a suite of benchmark but is dominated by some recent proposals that take advantage of the graph features.
Given the vast range of architectures, activation functions, regularizers, and optimization techniques that could be used within the NNMF framework, it seems likely the true potential of the approach has yet to be reached.
\end{abstract}

\section{Introduction}

We are interested in modeling arrays of data, which arise in the analysis of networks, graphs, and, more generally, relational data.  
For example,
in collaborative filtering and recommender system applications \citep{GNOT92,BiasedMF}, we may have $N$ users and $M$ movies, and for some collection $J \subseteq [N] \times [M]$ of user--movie pairs $(n,m)$, we have recorded the rating $X_{n,m}$ that user $n$ gave movie $m$.  The inferential goal we focus on in this work is to predict the ratings for those pairs not in $J$.  The data can be modeled as a partial observation of an $N \times M$ array $\bm X=(X_{n,m})$.  
While the methods we discuss are applicable far beyond the setting of movie rating data or even two-dimensional arrays, we will rely on the user--movie metaphor throughout.

One of the most popular approaches to modeling relational data using latent features is based on matrix factorization. Here the idea is 
to assume that $\bm X$ is well approximated by the product $\bm U^T \bm V$ of two low-rank matrices $\bm U \in \Reals^{D \times N}$ and $\bm V \in \Reals^{D \times M}$, where the rank $D$ is much less than $N$ and $M$.
Write $U_n$ and $V_m$ for the $D$-dimensional column vectors of $\bm U$ and $\bm V$, respectively.
Informally, we will think of $U_n$ as a latent vector of features describing user $n$, and of $V_m$ as a latent vector of features describing movie $m$.  The ranking $X_{n,m}$ is then approximated by the inner product $U_n ^T V_m$.
 
Probabilistic matrix factorization \citep{RussPMF}, or simply PMF, is based on matrix factorization, and further assumes the entries of $\bm X$ are independent Gaussians with common variance and means given by the corresponding entries of $\bm U^T \bm V$.  Maximum likelihood inference of the features $\bm U$ and $\bm V$ leads one to minimize the Frobenius norm of $\bm X - \bm U^t \bm V$, or equivalently, the root mean squared error (RMSE) between the inner product of the features and the observed relations.  In practice, regularization of the features vectors often improves the performance of the resulting predictions, provided the regularization parameter is chosen carefully, e.g., by cross validation.
 
 PMF is extremely effective in practice, but is also easy to improve upon when dealing with large but sparsely observed array data, as is typical in collaborative filtering. 
 One way to improve upon PMF is to introduce row and column effects that model systematic biases associated with users and with movies, leading to a model known in collaborative filtering community as BiasedMF \citep{BiasedMF}.  In this approach, the mean of $X_{n,m}$ is taken to be $U_n^T V_m + \mu_n + \tau_m + \beta$, where $\mu = (\mu_1,\dotsc,\mu_S)$, $\tau=(\tau_1,\dotsc,\tau_M)$, and $\beta$ are additional latent variables representing the user, movie, and global biases, respectively.   Note that the row and column effects in BiasedMF can be seen as a special case of PMF where we fix an entry of $\bm U$ and a  distinct entry of $\bm V$ to take the value~1.  In other words, BiasedMF implements a strong inductive bias.
Again, regularization improves prediction performance in practice.

In this short paper, we describe a different approach to factorizing $X$.  
Write $U_n \hprod V_m$ for the element-wise product, i.e., the $D$-dimensional vector whose $i$'th entry is $U_{i,n} V_{i,m}$. 
Using this notation, PMF models the mean of $X_{n,m}$ by $f(U_n \hprod V_m)$, where $f$ is the function $f(w_1,w_2,\dotsc) = \sum_j w_j$.  In the same notation, BiasedPMF models the mean of $X_{n,m}$ by $f(U_n \hprod V_m, \mu_n, \tau_m, \beta)$.
Our idea is to learn $f$, rather than assume it is fixed.  In particular, we take $f=f_\theta$ to be a feed-forward neural network with weights $\theta$.  Given data, we learn the weights $\theta$ at the same time that we learn the latent features.  Note that, for fixed latent feature vectors, we effectively have a supervised learning problem, and so we can optimize the neural network by gradient descent as is typical.  Fixing the neural network, we can optimize the latent feature vectors also by gradient descent, not unlike recent applications of neural networks to the problem of transferring artistic styles to ordinary images \citep{NeuralArt}.
  As one would expect, regularization is critical.  We used $\ell_2$-regularization for the latent feature vectors, and chose the regularization parameter $\lambda$ by optimizing the error on a validation set. %
  We call our proposal \emph{neural network matrix factorization}, or simply NNMF.

\section{Model}
\label{model}

Let $[N] = \{1,2,\dotsc,N\}$.
A data array is modeled as a collection of real-valued random variables
$X_{n,m}$, for $(n,m) \in J$, where $J \subset [N] \times [M]$ are the indices of the observed entries of the $N \times M$ data array.
(The extension to higher-dimensional arrays or arrays whose entries are elements in spaces other than $\Reals$ is straightforward.)

\newcommand{\DV}{D}
\newcommand{\DM}{{D'}}
\newcommand{\uu}{U}
\newcommand{\UU}{U'}
\newcommand{\vv}{V}
\newcommand{\VV}{V'}

To each row, $n \in N$, 
we associate a 
latent feature vector $\uu_n \in \Reals^\DV$
and 
a latent feature \emph{matrix} $\UU_n \in \Reals^{\DM \times K}$.
Similarly, 
to each column, $m \in M$, we associate a 
latent feature vector $\vv_m \in \Reals^{\DV}$
and 
latent feature matrix $\VV_m \in \Reals^{\DM \times K}$.
Write $\UU_{n,k}$ for the $k$'th column of $\UU_n$ (and similarly for $\VV_m$), and write $(U,V)$ for the collection of all latent features.\footnote{Note that there is nothing forcing the latent feature vectors $\uu_n$ for users and $\vv_m$ for movies to have the same dimensionality.  We made this choice for simplicity.}

Let $f_\theta$ be a feed-forward neural network with weights $\theta$. 
Viewing $\theta$ and the latent features $(U,V)$ as unknown parameters,
we assume the entries $X_{n,m}$ are independent random variables with means
\[
\hat X_{n,m} \defas 
\hat X(\uu_n,\vv_m,\UU_n,\VV_m) 
\defas f_\theta \bigl(\uu_n, \vv_m,\UU_{n,1}\hprod \VV_{m,1},\dotsc,\UU_{n,\DM} \hprod \VV_{m,\DM}\bigr).
\]
In other words, the neural network $f_\theta$ has $2\DV+\DM$ real-valued input units and one real-valued output unit.
The first $\DV$ are user-specific features; the next $\DV$ are movie-specific features;
and the last $\DM$ inputs are the result of inner products between $K$-dimensional vectors.\footnote{In our experiments, we took $K=1$ and $\DM$ large.  Taking $\DM=1$ results in a model where the input to the neural network is the prediction of a rank-$K$ matrix factorization and $2\DV$ additional features.  This simple modification of matrix factorization lead to improvements, but not as dramatic as those we report in \cref{experiments}.}

\section{Learning}
\label{learning}

To learn the network weights $\theta$ and latent features $(U,V)$,
we minimize the objective
\[\label{objfunc}
 \sum_{(n,m) \in J} (X_{n,m} - \hat X_{n,m})^2 
 +  \lambda \Bigl [
      \sum_n ||\UU_n||^2_{F} 
 +  \sum_n ||\uu_n||^2_{2} 
 +  \sum_m ||\VV_m||^2_{F} 
 +  \sum_m ||\vv_m||^2_{2} \Bigr ],
\]
where $\lambda$ is a regularization parameter,
$||\cdot||_2$ denotes the $\ell_2$ norm, and  $||\cdot ||_{F}$ denotes the Frobenius norm.
This objective can be understood as a penalized log likelihood under a Gaussian model for each entry.  
It can also be understood as a specifying the maximum \emph{a posteriori} estimate assuming independent Gaussian priors for every latent feature.

During training, we alternated between optimizing the neural network weights, while fixing the latent features, 
and optimizing the latent features, while fixing the network weights. 
Optimization was carried out by gradient descent on the entire dataset (i.e., we did not use batches). We used RMSProp to adjust the learning rate.  (See \cref{experiments} for details.) We did not evaluate other optimization algorithms.

\section{Related Work}

NNMF is very similar in spirit to the Random Function Model for modeling arrays/matrices proposed by \citet{RFMLloyd}.  Using probabilistic symmetry considerations, they arrive at a model where the mean of $X_{n,m}$ is given by $g(U_n,V_m)$, where $g : \Reals^{2D} \to \Reals$ is modeled by a Gaussian process.
At a  high level, our model replaces the Gaussian process prior with a parametric neural network one. 
We explored simply taking $g$ to be a feed-forward neural network (acting on a concatenated pair of vectors in $\Reals^D$), but found that we achieved better performance if some of the input dimensions first underwent an element-wise product. (We discuss this further below in relationship to NTN models.)  Conceivably, a deep neural network could learn to approximate the element-wise product or even outperform it, but this was not the case in our experiments, which used gradient-descent techniques to learn the neural network weights. In experiments, we found that NNMF significantly outperformed RFM, although the  RFM results were those produced by an implementation that was limited to $D \le 3$ latent dimensions due to significant algorithmic issues associated with scaling up inference for the Gaussian process component.  Given some recent advances in GP inference, it would be interesting to revisit the RFM, though it is not clear to the authors whether the advances are quite enough.

NNMF is related to some methods applied to Knowledge Bases and Knowledge Graphs. (See \citep{NickelMurphy2015} for a review of relational machine learning and Knowledge Graphs.)
Knowledge bases (KBs) are relational data composed of entity--relation--entity triples.   
For example, a geopolitical knowledge base might contain facts such as
\texttt{(Rome, capitol-of, Italy)} and \texttt{(Lithuania, member-of, EU)}.
Knowledge graphs (KGs) are representations of KBs as graphs
whose vertices represent entities and whose (labelled) edges represent relations between entities. 
Given this connection, one can see that KBs can be thought of as a collection of (extremely sparsely observed) arrays, one for each relation, or as a single three-dimensional array.
The key challenge in modeling KBs and KGs is to simultaneously learn many relations using shared representations in order to augment the limited data one has on each relation.  This is one of the key differences with the collaborative filtering setting. 

A method for KGs similar to NNMF is the Neural Tensor Network (NTN), which combines a tensor product with a single-layer neural network~\citep{NTN2013}.  
(Methods similar to NTN have been applied to problems in speech recognition. See, e.g., \citep{YuDS2013}.)
Other approaches to KGs use neural networks to produce representations, rather than map representations to predictions, like NTN and NNMF.  (See, e.g., \citep{HuangHJ15} and \citep{BGL2014}.)

NTNs model each element of a two-dimensional array by
\[\label{tensorrep}
\hat X_{n,m}  
\defas a^T \tanh \bigl( U_n^T Q^{[1:H]} V_m +  W
     \begin{bmatrix}
        U_n\\ %
        V_m %
     \end{bmatrix}
 	+ b\bigr),
\] 
where 
$U_n, V_m \in \Reals^D$  are feature vectors; 
$a \in \Reals^H$ is a linear layer weight vector; 
$b \in \Reals^H$ is a bias vector;
$W \in \Reals^{H \times 2D}$ is a weight matrix;
$Q^{[1:H]} \in \Reals^{D \times D \times H}$ is a third order tensor;
and the nonlinearity $\tanh (\cdot) : \Reals^H \to \Reals^H$ acts element wise.
 The tensor product term $U_n^T Q^{[1:H]} V_m$ denotes the element in $\Reals^H$ whose $h$'th entry is equal to $U_n^T Q^h V_m $, where $Q^h \in \Reals^{D \times D}$ is the $h$'th slice of $Q^{[1:H]}$. The model is trained by optimizing the \emph{contrastive max-margin} objective using L-BFGS with mini-batches.
 
Ignoring the particularly nonlinearity used,
the \emph{first} layer of the NNMF model can be expressed in the form \cref{tensorrep} if we take $W=0$ and allow ourselves to fix some entries of the latent features. (NTN employs no additional layers.)
For example, taking $K=1$ as in our experiments, 
define 
for $n \in [N]$, $m \in [M]$, $i,j \in [2\DV+\DM]$, and $h \in [H]$,
\[
\bar U_n = 
 \begin{bmatrix}
        \uu_n \\
        1_\DV  \\
        \UU_n 
     \end{bmatrix}  \in \Reals^{2\DV+\DM}, 
     \qquad    
\bar V_m=
  \begin{bmatrix}
        1_\DV \\ 
        \vv_m \\
        \VV_m 
     \end{bmatrix} \in \Reals^{2\DV+\DM}, 
  \qquad
 Q_{ij}^h = \begin{cases}
      W'_{h,i}, & \text{if } i=j, \\
      0, &\text{otherwise,}     
\end{cases}
     \]
where $1_\DV$ denotes an $\DV$-dimensional column vector with all entries equal to 1
and $W' \in \Reals^{H \times (2\DV+\DM)}$ is the weight matrix defining the first layer of the NNMF network.
 Then we recover the first layer of NNMF with the third-order tensor 
 $Q^{[1:H]}  \in \Reals^{(2\DV+\DM) \times (2\DV+\DM) \times H}$ 
whose $h$'th slice is $Q^h$.

There have been many scalable techniques proposed to model very large KGs.
\citet{NickelMurphy2015} split existing models into two categories: latent feature models and graph feature models. Latent variable methods learn unobserved representations of entities and use them to predict relations,
while graph feature methods learn to predict relations directly from extracted features of local graph structure.
\citet{TC2015} argue through empirical comparisons that these two categories of models exhibit complimentary strengths.

A number of state-of-the-art proposals for collaborative filtering are perhaps best thought of as incorporating aspects of  graph feature models.
An example of a method relaxing the low-rank assumption using graph features is the Local Low Rank Matrix Approximation \citep{LLORMA}, which assumes that every entry in the matrix is given by a combination of low rank matrices, where the combination is specific to the entry.
LLORMA achieves impressive state-of-the-art performance.

Other approaches also use neural-network architectures but work by trying to predict the ground truth ratings directly from the observed ratings matrix $X$.  For example,
in I-AutoRec \citep{AutoRec}, an autoencoder is learned that takes as input the observed movie ratings vector $X_n$ for user $n$ and produces as output the ground truth $X^{\textrm{truth}}_n$. (Missing entries are typically replaced by value~3.)
AutoRec achieves state-of-the-art performance, slightly besting LLORMA on some benchmarks, but a careful comparison would likely require a fresh data set and strict controls on how the numerous parameters for both models are chosen \citep{BH15,Dwork07082015}.
Another model in this category is the I-RBM \citep{I-RBM}, but its performance is now far from the state of the art.

Both LLORMA and I-AutoRec can be seen as models combining aspects of both graph feature and latent feature models. 
LLORMA identifies similar rows and columns (entities) using graph features, but model each local low-rank approximation using latent features. 
I-AutoRec takes as input all observed ratings (relations) for a user (entity), allowing the network to model the graph features, which in this case are similarities and distances among movies.

In \cref{experiments}, we compare the performance of NNMF and other approaches on benchmarks including link prediction in graphs, as well as collaborative filtering in movie rating datasets.
In our experiments, NNMF dominated other latent feature methods, as well as the I-RBM model.  However, NNMF was dominated by both LLORMA and I-AutoRec.
One possibility is that a different approach to learning the underlying neural network would deliver results on par with these methods. 
Another possibility is that the difference reflects some fundamental limitation of latent feature models, which assume that the ratings are conditionally independent given the latent feature representations. 
Local graph structure may contain information that would aid in predicting ratings.
In particular, NNMF does not learn from the pattern of missing ratings,which can reveal information about a user or movie: e.g., a user might tend only to give ratings when those ratings are extreme, and movies with low ratings are less likely to be viewed in the first place.
In contrast to NNMF, both LLORMA and AutoRec could, in principle, be taking advantage of the information latent in the pattern of missing ratings, although the strength of this effect has not been studied.  In LLORMA, the sparsity pattern affects the notion of locality.  In AutoRec, the entire pattern of ratings is fed as input, although the sparsity is obscured somewhat by missing entries being replaced by 3's.

Some recent work by \citet{LHG14} demonstrates that explicitly modeling the non-random pattern of missing ratings can lead to a slight improvement in performance for latent feature models, although the gains they demonstrated were not dramatic enough that they would have closed the gap between NNMF and LLORMA/AutoRec.
Indeed, we implemented a neural architecture similar in spirit to theirs, but were only able to improve the RMSE score by approximately $0.003$.   A more careful analysis would be necessary to make more definitive conclusions. 

\begin{table}[t]
\centering
\begin{tabular}{lllll}
           & NIPS  & Protein & ML100k & ML1m    \\
&           \\
Vertices X & 234   & 230     & 943    & 6040    \\
Vertices Y & -   & -       & 1682   & 3900    \\
Edges      & 27144 &   52900      & 100000 & 1000209
\end{tabular}
\caption{Data sets and their dimensions.  The mark ``-'' highlights that the array is square.}
\label{datasetsinfo}
\end{table}

\begin{table}[t]
\centering
\begin{tabular}{lllll}
              & NIPS  & Protein & ML-100K \\
              & \\
RFM (3)          & 0.110 & 0.136   & -           \\  %
PMF (3)          &  0.130 &  0.139   & -     \\  %
PMF (60)          & 0.062 & 0.104   & 0.952    \\  %
BiasedMF  (60)     &  0.065    & 0.111        & 0.911  \\   %
NTN (60)               & 0.048 & 0.071 & 0.910 \\ %
 \vspace*{-.75em}& \\      
NNMF (3HL)          & 0.040 & 0.065   & 0.907 \\    %
NNMF (4HL)          & - & -  & 0.903    %
\\ & 
\\ & \\
\end{tabular}\ \ \ \ 
\begin{tabular}{lllll}
              & ML-1M \\
              & \\
PMF (60)               &  0.883 \\ %
LLORMA-GLOBAL    & 0.865 \\  %
I-RBM                        &0.854 \\   %
BiasedMF (60)     & 0.852 \\  %
NTN (60)                   & 0.852  \\ %
LLORMA-LOCAL       & 0.833 \\  %
I-AutoRec                  & 0.831 \\  %
 \vspace*{-.75em}& \\      
NNMF (3HL)                        & 0.846   \\ %
NNMF (4HL)			& 0.843
\end{tabular}
\caption{Results across the four data sets for a variety of techniques.  
The token $(D)$ specifies that a rank-$D$ factorization was used. 
The token $(n \mathrm{HL})$ specifies that $n$ hidden layers were used. Scores reported for RFM and PMF (3) are taken from \citep{RFMLloyd}.  Scores for BiasedMF were obtained using LibRec \citep{LibRec}. Scores for LLORMA were taken from  \citep{LLORMA}, AutoRec and RBM, were taken from \citep{AutoRec}.}
\label{resultstable}
\end{table}

\section{Experiments}
\label{experiments}

We evaluated NNMF on two graph datasets (NIPS and Protein) and two collaborative filtering datasets (MovieLens 100K and 1M).  See \cref{datasetsinfo} for more information about the datasets.

NNMF, NTN, PMF model performance was evaluated on 5 randomly subsampled test sets, each comprising $10\%$ of the data points, and then averaged. 
The remaining $90 \%$ of the data was split into training and validation sets:
For the graph datasets, we used a $10\%$ of the training data for validation.
Due to the larger size of collaborative filtering datasets, we used $2\%$ and $0.5\%$ of the training data for validation on the MovieLens 100K and 1M datasets, respectively. These numbers were chosen to make the Monte Carlo error of the validation set estimate sufficiently small. \NA{(It is likely that results could be improved by better use of the training and validation data.)}

The regularization parameter, $\lambda$, and optimal stopping time were chosen by optimizing the error on the validation set.  For every fixed setting of $\lambda$, the network and features were learned by optimizing
\cref{objfunc} as described in \cref{learning}.

For simplicity, and to avoid the pitfalls of choosing parameters that produce good test set performance, 
the number and dimensionality of the features, as well as the network architecture, were fixed across experiments. It is conceivable that cross validating these parameters would have yielded better results.  
On the other hand, it would be wise to employ safeguards \citep{Dwork07082015} before embarking on an adaptive search for better architectures, learning rates, activation functions, etc.

We chose $\DM=60$ feature dimensions to be preprocessed by an element-wise product, and included $\DV=10$ additional features for each user and each movie. 
The feed-forward neural network was chosen to have 3 hidden layers with 50 sigmoidal units each.
The network weights were sampled uniformly in $\pm 4 \sqrt {6} /  \sqrt {n_{\mathrm{in}} + n_{\mathrm{out}}}$, where
$n_{\mathrm{in}}$, $n_{\mathrm{out}}$ denote the number of inbound and outbound connections.
The latent features were randomly initialized from a zero-mean Gaussian distribution with standard deviation $0.1$.  
The features and weights were learned by gradient descent, and RMSPROP was used to adjust the learning rate, which was initialized to $0.001$ for NIPS, Protein, and ML-100K, and to $0.005$ for ML-1M.

To train the PMF model, we chose 60 dimensions after evaluating the performance of PMF with various choices for the dimensionality and finding that this worked best.  On each run, the regularization parameter was chosen from a large range by optimizing the validation error.  (We tried many other settings for PMF, and have reported the best numbers we obtained here to make the comparison conservative.) 

NTN model hyperparameters were chosen to match NNMF ones---we used 60-dimensional latent features, and $50$ units in the hidden layer. This setup yields a third order tensor with $60 \times 60 \times 50 = 180,000$ entries. Compared to the network underlying NNMF, a NTN of approximately the same size has roughly 20 times more parameters. The model was trained with gradient descent on the same objective function as for NNMF. We had to use mini-batches for the MovieLens 1M dataset to avoid memory issues. Just as for other models, we chose the regularization parameter $\lambda$ by optimizing it the error on the validation set. Note, that in the original NTN model was trained with a contrastive max-margin objective function with $\ell_2$ regularization of all parameters. We applied a sigmoid nonlinearity to the output layer of the original NTN, to ensure that its outputs fell in $[0,1]$.

The results appear in \cref{resultstable}.  As mentioned above, NNMF dominates PMF, RFM, and to a lesser extent NTN.  In \citep{RFMLloyd}, the performance of RFM is compared with PMF when both models use the same number of latent dimensions.  The performance of PMF, however, tends to improve with the higher dimension, assuming proper regularization, and so RFM (3) is seen here to perform worse than PMF (60).  It is possible that recent advances in Gaussian process regression could in turn improve the performance of RFM.   

NNMF outperforms BaisedMF, although the margin narrows as we move to the sparsely-observed MovieLens datasets. We note that adding bias correction terms to NNMF also improves the performance of NNMF, although the improvement is on the order of $0.003$, and so may not be robust. It is also possible that using more of the training data might widen the gap.  

NNMF beats the (low-rank) global version of LLORMA, but not the local version that relaxes the low-rank constraint.  NNMF is also bested by AutoRec.    It is also not clear if we could have reliably found much better network weights and features had we made different choices around the architecture, composition, and training of the neural network.
Given that NNMF dominates PMF so handily on the graph datasets, it might stand to reason that there is a lot of room for improvement on MovieLens through better engineering of NNMF.  
It is worth noting that a `local' versions of NNMF could be developed along the same lines as were for LLORMA. Given that NNMF dominates PMF, it might then also stand to reason that a local version of NNMF would dominate LLORMA, because LLORMA can be understood as a local version of PMF.

To see whether deeper networks performed better on the collaborative filtering datasets, we also evaluated NNMF on the MovieLens data sets using a 4 hidden layer network.  We observed that fewer units per layer yielded better results. (We compared 50 units per layer when $(\DV,\DM)=(10,60)$ to 20 units per layer when $(\DV,\DM) = (10,80)$.) However, to draw any conclusions, more experiments would be needed, with care to avoid overfitting. We reported scores for 4 hidden layer networks, with 20 units per hidden layer, and $(\DV,\DM)=(10,80)$ latent feature dimensions.
We believe that adding additional layers would likely improve the results, though we suspect the performance would saturate quickly (and then drop if we did not very carefully initialize and regularize the network).

\section{Discussion}

NNMF achieves state-of-the-art results among latent feature models, but is dominated by approaches that take into account local graph structure.  However, it is possible that our experiments have not identified the limits of the NNMF model.  It is difficult to exhaustively explore the range of network architectures, activation functions, regularization techniques, and cross-validation strategies.  Even if we could explore them all, we would be in danger of overfitting and losing any hope of insight into the usefulness of NNMF.  Indeed, we erred towards not trying to optimize over the model's many possible configuration.  It would be interesting to apply recent advances in adaptive estimation to control the possibility of overfitting during this phase of designing and evaluating a new model \citep{Dwork07082015}.

\section*{Acknowledgments}

The authors would like to thank Zoubin Ghahramani for feedback and helpful discussions.

\bibliographystyle{abbrvnat}
\bibliography{biblio}

\begin{thebibliography}{18}
\providecommand{\natexlab}[1]{#1}
\providecommand{\url}[1]{\texttt{#1}}
\expandafter\ifx\csname urlstyle\endcsname\relax
  \providecommand{\doi}[1]{doi: #1}\else
  \providecommand{\doi}{doi: \begingroup \urlstyle{rm}\Url}\fi

\bibitem[Bian et~al.(2014)Bian, Gao, and Liu]{BGL2014}
J.~Bian, B.~Gao, and T.-Y. Liu.
\newblock Knowledge-powered deep learning for word embedding.
\newblock In \emph{Proc. European Conference on Machine Learning}. Springer,
  September 2014.

\bibitem[Blum and Hardt(2015)]{BH15}
A.~Blum and M.~Hardt.
\newblock The ladder: A reliable leaderboard for machine learning competitions,
  2015.
\newblock arXiv:1502.04585. Conference version appeared in ICML, 2015.

\bibitem[Dwork et~al.(2015)Dwork, Feldman, Hardt, Pitassi, Reingold, and
  Roth]{Dwork07082015}
C.~Dwork, V.~Feldman, M.~Hardt, T.~Pitassi, O.~Reingold, and A.~Roth.
\newblock The reusable holdout: Preserving validity in adaptive data analysis.
\newblock \emph{Science}, 349\penalty0 (6248):\penalty0 636--638, 2015.
\newblock Earlier versions appeared in NIPS 2015 and STOC 2015.

\bibitem[Gatys et~al.(2015)Gatys, Ecker, and Bethge]{NeuralArt}
L.~A. Gatys, A.~S. Ecker, and M.~Bethge.
\newblock A neural algorithm of artistic style, 2015.
\newblock arXiv:1508.06576.

\bibitem[Goldberg et~al.(1992)Goldberg, Nichols, Oki, and Terry]{GNOT92}
D.~Goldberg, D.~Nichols, B.~M. Oki, and D.~Terry.
\newblock Using collaborative filtering to weave an information tapestry.
\newblock \emph{Communications of the ACM}, 35:\penalty0 61--70, 1992.

\bibitem[Guo et~al.(2015)Guo, Zhang, Sun, and Yorke-Smith]{LibRec}
G.~Guo, J.~Zhang, Z.~Sun, and N.~Yorke-Smith.
\newblock Librec: A java library for recommender systems, 2015.
\newblock In Posters, Demos, Late-breaking Results and Workshop Proceedings of
  the 23rd Conference on User Modelling, Adaptation and Personalization (UMAP).

\bibitem[Hern\'andez-Lobato et~al.(2014)Hern\'andez-Lobato, Houlsby, and
  Ghahramani]{LHG14}
J.~M. Hern\'andez-Lobato, N.~Houlsby, and Z.~Ghahramani.
\newblock Probabilistic matrix factorization with non-random missing data.
\newblock In \emph{Proc. of the Int. Conf. on Machine Learning}, 2014.

\bibitem[Huang et~al.(2015)Huang, Heck, and Ji]{HuangHJ15}
H.~Huang, L.~Heck, and H.~Ji.
\newblock Leveraging deep neural networks and knowledge graphs for entity
  disambiguation, 2015.
\newblock arXiv:1504.07678v1.

\bibitem[Koren et~al.(2009)Koren, Bell, and Volinsky]{BiasedMF}
Y.~Koren, R.~Bell, and C.~Volinsky.
\newblock Matrix factorization techniques for recommender systems.
\newblock \emph{Computer}, 42\penalty0 (8):\penalty0 30--37, Aug. 2009.
\newblock \doi{10.1109/MC.2009.263}.

\bibitem[Lee et~al.(2013)Lee, Kim, Lebanon, and Singer]{LLORMA}
J.~Lee, S.~Kim, G.~Lebanon, and Y.~Singer.
\newblock Local low-rank matrix approximation.
\newblock In \emph{Proc. of the Int. Conf. on Machine Learning}, 2013.

\bibitem[Lloyd et~al.(2012)Lloyd, Orbanz, Ghahramani, and Roy]{RFMLloyd}
J.~Lloyd, P.~Orbanz, Z.~Ghahramani, and D.~M. Roy.
\newblock Random function priors for exchangeable arrays with applications to
  graphs and relational data.
\newblock In P.~Bartlett, F.~Pereira, C.~Burges, L.~Bottou, and K.~Weinberger,
  editors, \emph{Adv Neural Inform. Proc. Systems 25}, pages 1007--1015, 2012.

\bibitem[Nickel et~al.(2015)Nickel, Murphy, Tresp, and
  Gabrilovich]{NickelMurphy2015}
M.~Nickel, K.~Murphy, V.~Tresp, and E.~Gabrilovich.
\newblock A review of relational machine learning for knowledge graphs, 2015.
\newblock arXiv:1503.00759.

\bibitem[Salakhutdinov and Mnih(2008)]{RussPMF}
R.~Salakhutdinov and A.~Mnih.
\newblock Probabilistic matrix factorization.
\newblock In \emph{Neural Information Processing Systems 21}, 2008.

\bibitem[Salakhutdinov et~al.(2007)Salakhutdinov, Mnih, and Hinton]{I-RBM}
R.~Salakhutdinov, A.~Mnih, and G.~Hinton.
\newblock Restricted boltzmann machines for collaborative filtering.
\newblock In \emph{Proceedings of the 24th International Conference on Machine
  Learning}, ICML '07, pages 791--798, New York, NY, USA, 2007. ACM.
\newblock \doi{10.1145/1273496.1273596}.

\bibitem[Sedhain et~al.(2015)Sedhain, Menon, Sanner, and Xie]{AutoRec}
S.~Sedhain, A.~K. Menon, S.~Sanner, and L.~Xie.
\newblock Autorec: Autoencoders meet collaborative filtering.
\newblock In \emph{Proceedings of the 24th International Conference on World
  Wide Web}, WWW '15 Companion, pages 111--112, Republic and Canton of Geneva,
  Switzerland, 2015. International World Wide Web Conferences Steering
  Committee.
\newblock \doi{10.1145/2740908.2742726}.

\bibitem[Socher et~al.(2013)Socher, Chen, Manning, and Ng]{NTN2013}
R.~Socher, D.~Chen, C.~D. Manning, and A.~Ng.
\newblock Reasoning with neural tensor networks for knowledge base completion.
\newblock In C.~Burges, L.~Bottou, M.~Welling, Z.~Ghahramani, and
  K.~Weinberger, editors, \emph{Advances in Neural Information Processing
  Systems 26}, pages 926--934, 2013.

\bibitem[Toutanova and Chen(2015)]{TC2015}
K.~Toutanova and D.~Chen.
\newblock Observed versus latent features for knowledge base and text
  inference.
\newblock In \emph{3rd Workshop on Continuous Vector Space Models and Their
  Compositionality}. Association for Computational Linguistics, July 2015.

\bibitem[Yu et~al.(2013)Yu, Deng, and Seide]{YuDS2013}
D.~Yu, L.~Deng, and F.~Seide.
\newblock The deep tensor neural network with applications to large vocabulary
  speech recognition.
\newblock \emph{IEEE Transactions on Audio, Speech, and Language Processing},
  21\penalty0 (2):\penalty0 388--396, 2013.

\end{thebibliography}

\vfill

\end{document}